\documentclass{article}
\usepackage{spconf,amsmath,graphicx, verbatim,xcolor,hyperref}
\usepackage{amsfonts}
\usepackage{subcaption}
\usepackage{cleveref}

\title{Learnings from Federated Learning in the Real world}
\name{Christophe Dupuy$\:^{1}$, Tanya G. Roosta$\:^{1}$, Leo Long$\:^{1}$, Clement Chung$\:^{1}$, Rahul Gupta$\:^{1}$, Salman Avestimehr$\:^{1, 2}$}
\address{$\:^{1}$Amazon Alexa AI, Cambridge, MA, USA; $\:^{2}$University of Southern California (USC), CA, USA}
\begin{document}
%
\maketitle
\begin{abstract}
Federated Learning (FL) applied to real world data may suffer from several idiosyncrasies. One such idiosyncrasy is the data distribution across devices. 
Data across devices could be distributed such that there are some ``heavy devices” with large amounts of data while there are many ``light users” with only a handful of data points. 
There also exists heterogeneity of data across devices.
In this study, we evaluate the impact of such idiosyncrasies on Natural Language Understanding (NLU) models trained using FL. 
We conduct experiments on data obtained from a large scale NLU system serving thousands of devices and show that simple non-uniform device selection based on the number of interactions at each round of FL training boosts the performance of the model. 
This benefit is further amplified in continual FL on consecutive time periods, where non-uniform sampling manages to swiftly catch up with FL methods using all data at once.
\end{abstract}
\section{Introduction}
\label{sec:intro}
Deep neural networks have shown great promise in a variety of application domains. The data required for training these models is often generated via devices interaction with edge devices. In a traditional training framework, the data is sent to a central server, where the model gets trained and periodically deployed on the edge devices. The desire to strengthen data privacy and security has driven a surge of interest in developing training methodologies that can operate on local data that is not transmitted to the cloud. A promising framework is Federated Learning (FL), where each device performs local training using its own data and only sends updated model parameters to the server. The server aggregates these updates to build the new model and sends the latest model back to each device.
In order to provide an FL solution that scales to millions of users, in most algorithms (e.g., \texttt{FedAVG} \cite{mcmahan2017fedavg}), the central server first selects a subset of devices uniformly at random, then computes a weighted average of their parameter updates to construct the global model. In real world scenarios with heterogeneous users, the contribution of different devices might significantly vary and uniform sampling might not provide the best strategy. As such, device selection for model aggregation is a critical component of FL algorithms.  

Most recent works that study device selection in FL (e.g., \cite{balakrishnan2021diverse,Nagalapatti_Narayanam_2021}) have focused on gradient-based methods, in which the local updates of the devices are compared to decide the selection and aggregation. These methods require the computations to be carried out across all devices in each round. In this paper, we investigate simpler device selection processes that are based on ``device activity'' (i.e. selecting devices based on their number of available training samples). Given the FL server already has access to this information, no additional computation/communication overhead is incurred. 
We also study temporal effects of training data in FL, and look at how different device selection and training strategies affect the performance of the model. Most FL papers present results on ``static'' datasets where the time dimension is ignored. \\
\textbf{Our contributions:} 
(1) we present the first study of simple device selection based on device activity applied to real-world data, exploring several sampling strategies and show that non-uniform sampling outperforms standard uniform sampling; 
(2) we study continual FL to investigate the realistic temporal aspects of FL and address limited data storage on-device. We demonstrate that non-uniform strategies combined with continual learning swiftly catches up with FL methods using all data at once (at the expense of longer delays between updates).


\section{Related Work}
\label{related-workd}
Standard FL algorithms such as \texttt{FedAVG} \cite{mcmahan2017fedavg} and \texttt{FedOpt} \cite{reddi2021fedopt}) have been widely studied. 
\texttt{FedAVG} has been proved to converge under a large set of assumptions (e.g., non- I.I.D. data distributions) \cite{yang2021achieving,ruan2021flexible, rizk2020dynamic, acar2021federated}. 
The performance of these algorithms has been evaluated in CV (e.g., \cite{mcmahan2017fedavg}), NLP (e.g., \cite{fedNLP}), and graph-structured data (e.g., \cite{fedGraphNN}).  
Recently, several works have looked at leveraging the results of previous local device update to compute the sampling probability for the next round. 
\cite{chen2020optimal} propose a probability based on the ``importance’' of each device, which is mostly determined by the norm of the latest local update. 
In \cite{cho2020client}, the authors show that sampling devices with higher losses leads to faster convergence compared to vanilla \texttt{FedAVG}. Similarly, other methods leverage the latest local gradients to update the device sampling probability \cite{balakrishnan2021diverse,Nagalapatti_Narayanam_2021}. These methods require all the devices to perform local updates which increases the communication overhead. Continual learning (e.g., \cite{Thrun95,Schwarz2018progress,hung2019compacting}) has become increasingly important for learning across a sequence of rounds and tasks.
More recently, continual learning has also been modeled in FL to study how one can leverage learned knowledge from different devices with different tasks~\cite{yoon2021continual}.
In this work, we focus on temporal aspects of continual FL, meaning leveraging the knowledge from past interactions in the future training.

\begin{figure*}[t]
\centering
\begin{subfigure}[t]{0.7\textwidth}
  \centering
   \includegraphics[width=\linewidth]{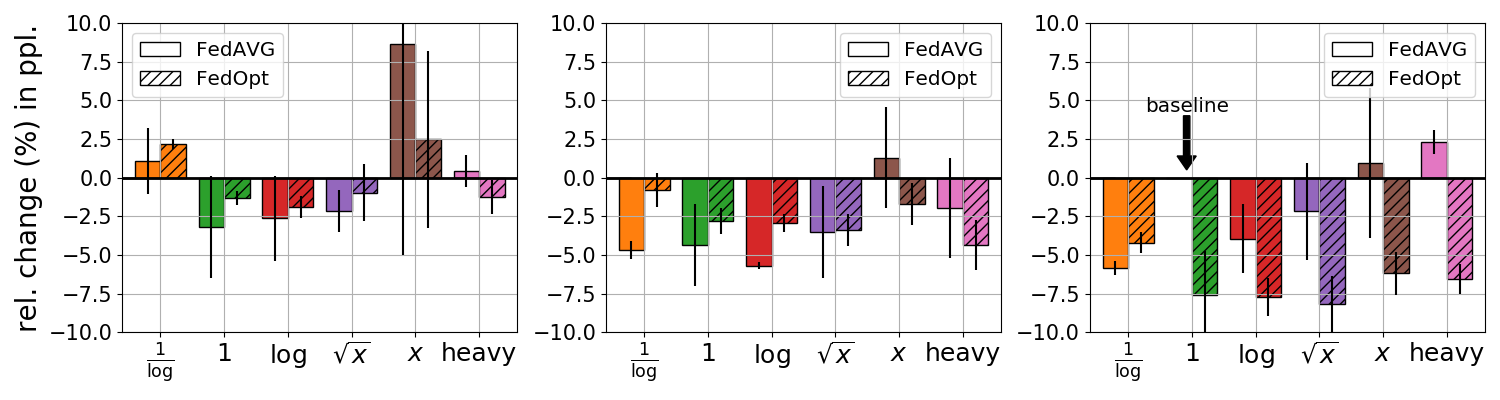}
	\caption{\textit{One-shot FL}. \textbf{Left}: 3 months; \textbf{middle}: 6 months; \textbf{right}: 11 months.} 
	\label{fig:one_shot_overall}
\end{subfigure}
\begin{subfigure}[t]{0.29\textwidth}
  \centering
   \includegraphics[width=\linewidth]{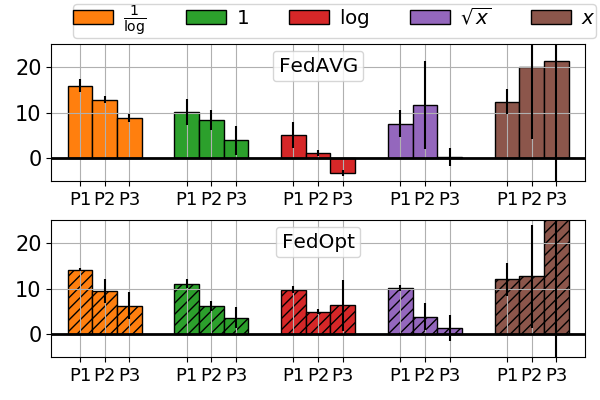}
	\caption{\textit{Continual FL}. Performance after each of the 3 time periods P1, P2, P3.}
	\label{fig:continual_overall}
\end{subfigure}
\caption{ 
RCP compared to the the baseline model on the whole test set.
(\ref{fig:one_shot_overall})~Baseline: standard \texttt{FedAVG}  ($f(x)=1$) on 11 months of data; (\ref{fig:continual_overall})~Baseline: \textit{one-shot FL} \texttt{FedAVG} and $f(x)=\log(x+1)$ on 6 months.}
\label{fig:overall}
\end{figure*}

\section{Problem Statement}
\label{sec:fl}
\textbf{Framework.} \hspace{2.5mm}
In standard machine learning, the central server stores the $n$ user generated entries $X=(x_i)_{i=1\ldots n}$ and the objective is to minimize the average loss $\mathcal{L}$ over these $n$ inputs:
$\min_{w} \mathcal{L}(X, w) = \frac{1}{n} \sum_{i=1}^{n} \ell(x_i, w)$.
In practice, the number of inputs $n$ is large and $\mathcal{L}$ is minimized with stochastic gradient descent (SGD) \cite{bottou2010sgd} where the gradient with respect to a random batch of inputs $B_t$ is used to update the parameters $w$:
$    w_{t+1} = w_{t} - \eta\nabla_w \left(\frac{1}{|B_t|}\sum_{i\in B_t} \ell(x_i, w)\right)$.
In FL, the data is not available to the central server to generate random batches of data $B_t$. If each of the $M$ devices contains $X^m\subset X$ inputs with $\vert X^m\vert=n_m$, $m=1,\ldots,M$, the central objective can be rewritten:
$\mathcal{L}(X, w) = \frac{1}{n} \sum_{m=1}^{M}n_m \mathcal{L}^m(X^m, w)$,
with $\mathcal{L}^m(X^m, w) =\frac{1}{n_m} \sum_{x\in X^m}\ell(x, w)$. In \texttt{FedAVG}, this objective is minimized by iteratively sampling a subset of devices $S_t\subset\{1, \ldots, M\}$, requiring the selected devices to optimize their own objective $\mathcal{L}^m(X^m, w_t)$ locally using the latest version of the model $w_t$ as the starting point. The updated local parameters $w^m_{t+1}$ are sent to the central server for aggregation via:
$w_{t+1} = \sum_{m\in S_t}\frac{n_m}{N_t} w^m_{t+1}$,
where $N_t=\sum_{m\in S_t}n_m$ is the total number of inputs on the selected devices $S_t$.

\noindent \textbf{Device Selection.} \hspace{2.5mm} 
In standard FL training, the device selection is made uniformly at random. In a real world scenario with non-I.I.D. data, different device selection strategies could help boost the performance of the resulting model. Previous works used gradient-based selection methods \cite{balakrishnan2021diverse, Nagalapatti_Narayanam_2021} where gradients are computed for all devices before selecting a subset. We look at a simpler/less computationally expensive approach using the number of inputs per device $(n_m)$. We want to understand: 1) change in performance if we put more weight on the active devices, 2) whether we can limit ourselves to only a subset of devices. In our experiments, we explore several selection strategies for $S_t$ that are a function of the number of inputs on each device, namely ${\mathbb{P}[m\in S_t]} \propto f(n_m)$,
with ${f(x)\in\{1, \log(x+1), \sqrt{x}, x, \frac{1}{ \log(x+1)}\}}$, where $f(x)=1$ corresponds to uniform sampling. We experiment with two additional functions referred to as \texttt{heavy} (resp. \texttt{light}) to restrict the optimization to the most (resp. least) active devices: $f(n_m) = 1 $ if $m$ is in top 20\% most (resp. least) used devices, $f(x)=0$ otherwise.

\noindent \textbf{Continual Learning.} \hspace{2.5mm} 
In the \textit{one-shot FL} setting, the data is static and used all at once during training. This approach might not be applicable to real-world scenario where users generate data over time due to limited storage on device. 
In \textit{continual FL}, we consider a dataset generated over a time period $T$, which is divided into consecutive segments of size $\Delta t$. For each segment, we train a model on the data for that segment, starting from the end model of the previous segment. For the first segment, we use the same starting point as for \textit{one-shot FL}.  Similar to \textit{one-shot}, we compare the different device selection strategies in a \textit{continual} setting. 

Our experiments address the following:
\textbf{Device selection:} 1) Is there a non-uniform sampling strategy that outperforms uniform sampling? 2) Can we obtain a similar performance as the standard uniform sampling by restricting training to the most active devices?
\textbf{Temporality:} 1) For \textit{one-shot FL}, does an increase in the amount of historical data have a positive effect on the performance on the next interactions? 2) Does \textit{continual FL} compete with \textit{one-shot FL}, and is there a device selection strategy that boosts \textit{continual FL} performance?

\section{Experimental Setting}
\label{sec:exp}
\textbf{Model:} BERT-\textit{base} model \cite{devlin2019bert} for the masked language modeling (MLM) task that is pre-trained model on the Common Crawl dataset \cite{commoncrawl}.  \\
\textbf{Data:} 12 months of automated transcription of Alexa user request data, that is de-identified to remove information that connects the data to the users. We randomly select 10,000 devices and their utterances. In all experiments, the test set corresponds to the last month of data. We repeat the process 3 times to generate different datasets and evaluate our approaches. In all the figures, the standard deviation across the 3 datasets is displayed.\\
\textbf{One-shot FL:} The FL models are trained on $\vert T\vert =3$ months (months 9 to 11 of the 12 months data) $\vert T\vert =6$ months (months 6 to 11) and $\vert T\vert =11$ months (months 1 to 11).\\
\textbf{Continual FL:} 6 months of data (months 6 to 11 of the whole data) is split into 3 parts of $\Delta t=2$ months. For each time period, the last state of the model from the previous period is used for initialization.\\
\textbf{FL methods:} We compare two of the most studied FL algorithms: \texttt{FedAVG} \cite{mcmahan2017fedavg} and \texttt{FedOpt} \cite{reddi2021fedopt}. For both, we select 800 devices per round and each device performs one local epoch over their data, with standard central aggregation.\\
\textbf{Evaluation:} We use the perplexity metric and study the impact across the different devices by splitting the test set into deciles of devices, based on their number of utterances. For each decile, the perplexity is computed on all the utterances by devices in that decile.
We can't present absolute performance values, instead we present relative change in \% in perplexity (referred to as RCP from this point on) compared to the perplexity on the entire test set for the baseline model. 

\begin{figure*}[t]
\begin{subfigure}[t]{0.74\textwidth}
  \centering
   \includegraphics[width=\linewidth]{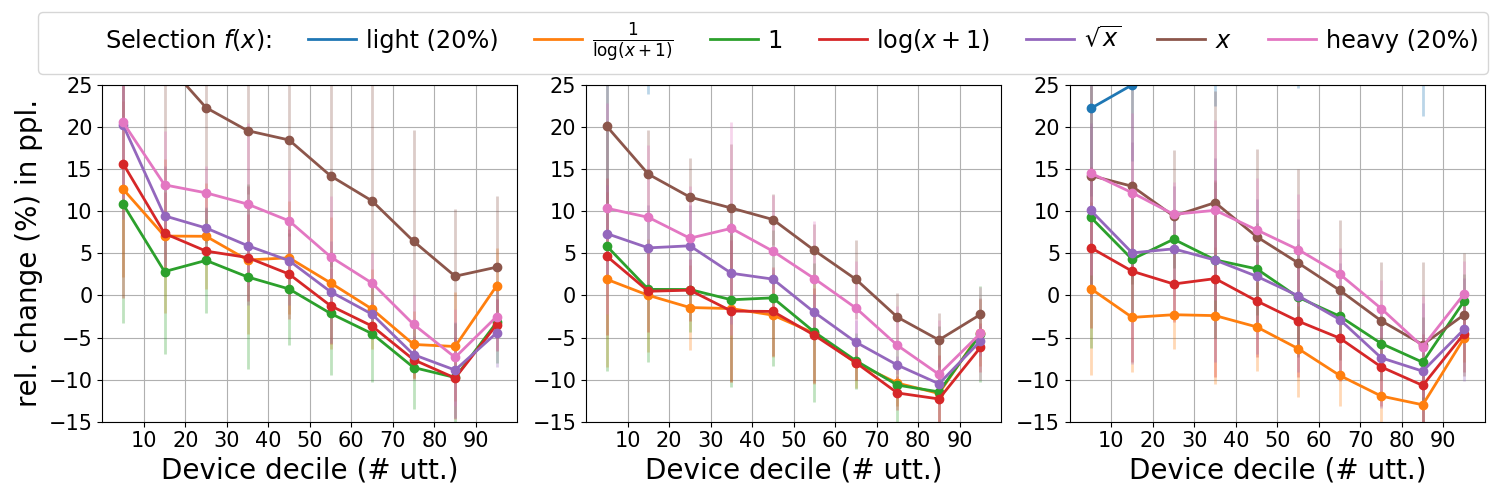}
	\caption{\textit{One-shot FL}. \textbf{Left}: 3 months; \textbf{middle}: 6 months; \textbf{right}: 11 months.}
	\label{fig:one_shot_fedavg}
\end{subfigure}
\begin{subfigure}[t]{0.24\textwidth}
  \centering
   \includegraphics[width=\linewidth]{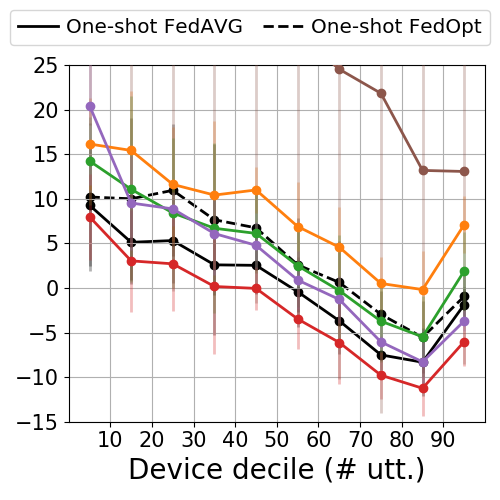}
	\caption{\textit{Continual FL}.}
	\label{fig:continual_fedavg}
\end{subfigure}
\caption{\texttt{FedAVG}. 
RCP compared to the the baseline model on the whole test set.
(\ref{fig:one_shot_fedavg})  Baseline: standard \texttt{FedAVG} on 11 months of data; (\ref{fig:continual_fedavg}) Baseline: \textit{one-shot FL} \texttt{FedAVG} and $f(x)=\log(x+1)$ on 6 months of data; \texttt{o-s} stands for ``\textit{one-shot}''.}
\label{fig:fedavg}
\end{figure*}

\begin{figure*}[t]
\begin{subfigure}[t]{0.74\textwidth}
  \centering
   \includegraphics[width=\linewidth]{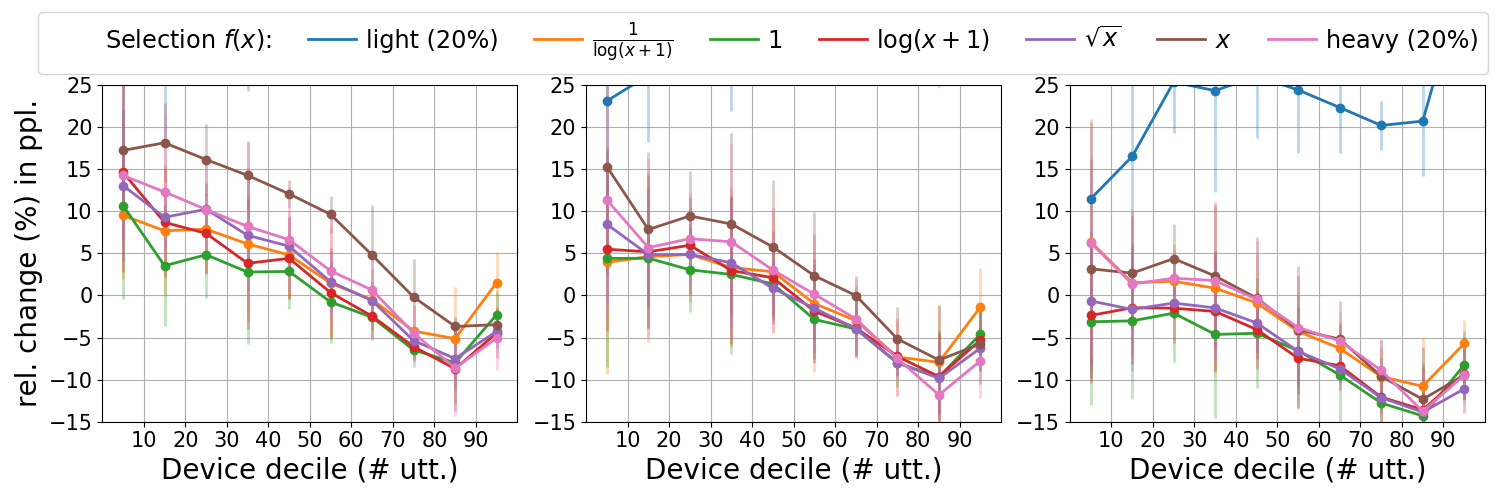}
	\caption{\textit{One-shot FL}. \textbf{Left}: 3 months; \textbf{middle}: 6 months; \textbf{right}: 11 months.}
	\label{fig:one_shot_fedopt}
\end{subfigure}
\begin{subfigure}[t]{0.24\textwidth}
  \centering
   \includegraphics[width=\linewidth]{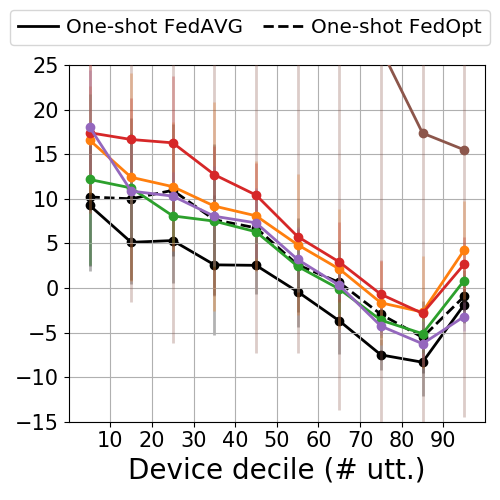}
	\caption{\textit{Continual FL}.}
	\label{fig:continual_fedopt}
\end{subfigure}
\caption{\texttt{FedOpt}. 
RCP compared to the the baseline model on the whole test set. 
(\ref{fig:one_shot_fedopt}) Baseline: standard \texttt{FedAVG} on 11 months of data; (\ref{fig:continual_fedopt}) Baseline: \textit{one-shot FL} \texttt{FedAVG} and $f(x)=\log(x+1)$ on 6 months of data; \texttt{o-s} stands for ``\textit{one-shot}''.}
\label{fig:fedopt}
\end{figure*}

\begin{figure}[t]
  \centering
  \includegraphics[width=\linewidth]{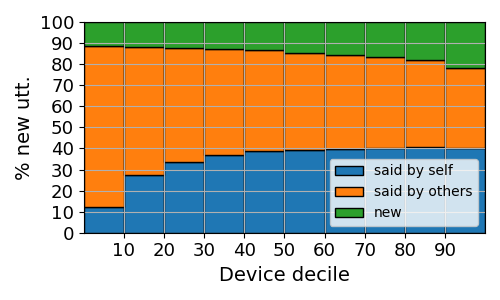}
  \caption{Average repartition of new utterances over a period.}
  \label{fig:innovation}
\end{figure}

\section{Results}
\noindent \textbf{One-shot FL  -- Figures~\labelcref{fig:one_shot_overall,fig:one_shot_fedavg,fig:one_shot_fedopt}.} \hspace{2mm}
The baseline for the RCP computation is the standard \textit{one-shot} \texttt{FedAVG} trained on 11 months of data.  
Increasing the amount of training data is clearly beneficial for \texttt{FedOpt}, but not as much for \texttt{FedAVG} (Figure~\ref{fig:one_shot_overall}). As a result, \texttt{FedOpt} outperforms \texttt{FedAVG} for most of the selection strategies with 11 months of training data. It's the other way around for $|T|\in\{3\text{ months}, 6\text{ months}\}$.
Interestingly, standard \texttt{FedAVG} using the entire 11 months of training data performs worse than almost any other strategy, possibly due to a combination of overfitting and sub-optimal selection strategy.
In Figures~\labelcref{fig:one_shot_fedavg,fig:one_shot_fedopt}, for every method, the perplexity is lower for devices with a higher number of utterances. Every curve sees an uptick in perplexity for the last decile. From the heaviest devices (especially the top 1\%), some also have a high number of utterances in the test set and are harder to predict.\\
\textbf{\texttt{FedAVG}}: the different sampling strategies significantly affect the performance of the resulting model. Overall, for shorter time range (left 2 plots in Figure~\ref{fig:one_shot_fedavg}), using the standard uniform sampling ($f(x)=1$) provides one of the best strategies, on par with the log-based sampling. 
However, for longer time horizon (right plot in Figure~\ref{fig:one_shot_fedavg}), non-uniform sampling provides much better results. The model where the probability to select device $m$ is proportional to $f(n_m)= \frac{1}{\log(n_m+1)}$ outperforms all the other models trained using \texttt{FedAVG}. In this strategy we tend to select lighter devices with higher probability.
Conversely, selection strategies that are restricted to a subset of devices (heavy or light devices only and $f(x)=x$) perform poorly. The selection $f(x)=x$ performs worse than when restricting sampling to the top 20\% devices because of the distribution of the devices: their number of utterances varies wildly across the dataset (with a factor of  $>10^4$), and choosing $f(x)=x$ implies the model is likely to select the same few devices (less than 20\% total) having much more utterances than the rest.  In terms of volume, it is not beneficial to extend the size of the data for \texttt{FedAVG}, except for the best strategy $f(x)=\frac{1}{\log(x+1)}$ that performs much better with 6 months over 3 months of data, but only slightly better when using 11 months compared to 6 months. In Figure~\ref{fig:one_shot_fedavg}, we see that more training data mostly benefits light devices. For instance, with 3 months of data and $f(x)=\frac{1}{\log(x+1)}$, the average RCP for the first decile is around 13\% and drops to around 1\% when using 11 months of data. For the last decile, RCP ``only'' drops from 2\% to -5\%.\\
\textbf{\texttt{FedOpt}}: contrary to \texttt{FedAVG}, the performance is similar across the selection strategies, except for selection restricted to \texttt{light} devices that performs poorly in general.  Restricting the training to the top 20\% devices fares as well as the other strategies involving all the devices. As expected, this strategy outperforms on heavier devices and underperforms on lighter devices which have not been seen by the model.
From Figure~\ref{fig:one_shot_fedopt}, uniform sampling provides better performance on light devices, but other strategies provide better performance on the overall test set ($f(x)\in\{\log(x+1), \sqrt{x}\}$). 

\noindent \textbf{Continual FL -- Figures~\labelcref{fig:continual_overall,fig:continual_fedavg,fig:continual_fedopt}.}  
The baseline is the best \textit{one-shot} model on the same 6 months: \texttt{FedAVG} trained with ${f(x)=\log(x+1)}$.
The advantage of \textit{continual FL} is the possibility to store less data on device but still train a model on the entire time period we consider.
From Figure~\ref{fig:continual_overall}, models trained with most selection strategies improve performance along the time periods, except $f(x)=x$.
Several selection strategies with \textit{continual FL} display less than 5\% relative degradation compared to the best \textit{one-shot FL} model after the third period of training (P3).  \textit{Continual FL} consistently outperforms the \textit{one-shot FL} baseline for \texttt{FedAVG} with $f(x)=\log(x+1)$. However, no other selection strategy beats the baseline with \textit{continual FL}, even if $f(x)=\sqrt{x}$ is on par with the baseline for both \texttt{FedAVG} and \texttt{FedOpt}.  Similar to \textit{one-shot FL}, the difference in performance across the selection strategies is larger for \texttt{FedAVG} than for \texttt{FedOpt}.

\textbf{\texttt{FedAVG}}: The gap between \textit{one-shot FL} and \textit{continual FL} across device deciles for $f(x)=\log(x+1)$ (black and red curves in Figure~\ref{fig:continual_fedavg}) indicates that \textit{continual FL} benefits the heavy devices. This is expected as heavier devices are more likely to have utterances in all the 3 periods used for training.

\textbf{\texttt{FedOpt}}: Even if all the selection strategies used with \texttt{FedOpt} perform worse than the \textit{one-shot} \texttt{FedAVG} baseline, they are consistently on par with the \textit{one-shot} \texttt{FedOpt} model across device deciles. Similar to \texttt{FedAVG}, \textit{continual FL} benefits heavy devices more. For instance,  \textit{continual FL} with \texttt{FedOpt} and $f(x)=\sqrt{x}$ performs worse than the \textit{one-shot} \texttt{FedOpt} baseline on the first device deciles, but outperforms this method on the last deciles.

\noindent \textbf{Insight.} \hspace{2mm} We look at the repartition of utterances for a given period compared to all the previous ones across device deciles. For a given device's set of unique utterances over a time period, we compute 1) the proportion of utterances already seen by this device; 2) the proportion of new utterances for this device but seen by any other device in a previous period of time; 3) the proportion of completely new utterances. In Figure~\ref{fig:innovation}, these metrics are averaged over periods P2 and P3. We observe that heavy devices explore more than light devices, who are mostly ``catching up''. For learning, it seems more beneficial to sample heavier devices who bring additional knowledge.

\section{Conclusion}
\label{sec:conclusion}
We studied popular FL algorithms on non-I.I.D. data borrowed from a real word setup. 
Through our experiments, we show that non-uniform device selection outperforms standard uniform sampling of FL and boosts performance for most devices.
We also compare the selection strategies on different amounts of data and observe that \texttt{FedOpt} benefits from more data, but not \texttt{FedAVG}. 
Finally, we compare \textit{one-shot FL} with \textit{continual FL} where we run multiple rounds of training on smaller time periods, allowing for lower storage capacity. 
We show that \textit{continual FL} outperforms the \textit{one-shot} strategy in some setting, and is overall most beneficial for heavy devices.


\vfill\pagebreak

\bibliographystyle{IEEEbib}
\bibliography{refs}

\end{document}